\documentclass{article}

\usepackage[final]{corl_2020} % Uncomment for the camera-ready ``final'' version.

\usepackage{xcolor}
\usepackage{amsmath,amsthm,amssymb}
\usepackage{graphicx}
\usepackage{comment}
\usepackage{subfigure}
\usepackage{array}
\newcolumntype{C}[1]{>{\centering\let\newline\\\arraybackslash\hspace{0pt}}m{#1}}

\usepackage{booktabs}
\usepackage{multirow}

\newcommand{\xxnote}[3]{}
\ifx\hidenotes\undefined
  \usepackage{color}
  \renewcommand{\xxnote}[3]{\color{#2}{#1: #3}}
\fi

\newcommand{\norm}[1]{\left\lVert #1 \right\rVert}
\DeclareMathOperator*{\argmax}{arg\,max}
\DeclareMathOperator*{\argmin}{arg\,min}

\newcommand{\tloss}[0]{\overline{\Delta}}
\newcommand{\ploss}[0]{\Delta}
\newcommand{\stateSpace}[0]{\mathcal{S}}
\newcommand{\actionSpace}[0]{\mathcal{A}}

\title{Learning Online from Corrective Feedback: A Meta-Algorithm for Robotics}

\author{
  Matt Schmittle\\
  Paul G. Allen School of Computer Science\\
  University of Washington 
  United States\\
  \texttt{schmttle@cs.washington.edu} \\
  \And
  Sanjiban Choudhury\\
  Paul G. Allen School of Computer Science\\
  University of Washington 
  United States\\
  \texttt{sanjibac@cs.washington.edu} \\
  \AND
  Siddhartha S.~Srinivasa\\
  Paul G. Allen School of Computer Science\\
  University of Washington 
  United States\\
  \texttt{siddh@cs.washington.edu} \\
}

\begin{document}
\maketitle

\begin{abstract}
A key challenge in Imitation Learning (IL) is that optimal state actions demonstrations are difficult for the teacher to provide. For example in robotics, providing kinesthetic demonstrations on a robotic manipulator requires the teacher to control multiple degrees of freedom at once. The difficulty of requiring optimal state action demonstrations limits the space of problems where the teacher can provide quality feedback. As an alternative to state action demonstrations, the teacher can provide corrective feedback such as their preferences or rewards. Prior work has created algorithms designed to learn from specific types of noisy feedback, but across teachers and tasks different forms of feedback may be required. Instead we propose that in order to learn from a diversity of scenarios we need to learn from a \emph{variety of feedback}. To learn from a variety of feedback we make the following insight: the teacher's cost function is latent and we can model a stream of feedback as a stream of loss functions. We then use any online learning algorithm to minimize the sum of these losses. With this insight we can learn from a diversity of feedback that is weakly correlated with the teacher's true cost function. We unify prior work into a general corrective feedback meta-algorithm and show that regardless of feedback we can obtain the same regret bounds. We demonstrate our approach by learning to perform a household navigation task on a robotic racecar platform. Our results show that our approach can learn quickly from a variety of noisy feedback.
\end{abstract}


\section{Introduction}
\label{introduction}

Imitation learning (IL) encompasses algorithms that leverage expert feedback to train dynamical systems. Key to its appeal is the observation~\citep{Bagnell-2015-5921,il_survey_now,Billard2012} that it is often easier for experts to provide insight via demonstrations or feedback rather than to hand-tune reward functions. Additionally, even when reward functions are pre-specified, IL demonstrates greater sample efficiency~\citep{deeplyaggrevated, il_survey_now} than pure reinforcement learning (RL), which is critical for robotics~\citep{motion_primitives, abbeel_heli} where real-world execution can be dangerous or expensive.

A key challenge to IL, however, lies in the details of how the expert interacts with the system. Most IL algorithms expect optimal state-action feedback. But, giving that feedback might be challenging, even for experts. For example, \citet{kinethetic} ran a user study on the viability of kinesthetic demonstrations using a manipulator for an assembly task and found minimal correlation between a person's performance at manual assembly (no robot) and their performance through kinesthetic teaching.

Prior work has addressed the difficulty in providing demonstrations by alternatively using corrective feedback. Corrective feedback does not require the teacher to be able to provide a full demonstration but instead adjustments during robot execution. For example, instead of providing kinesthetic demonstrations~\citet{Bajcsy:2017} simply physically pushes the robot during execution to the desired behavior. Even simplier, preference learning only asks the user to choose between two proposed actions~\cite{preference_learning}. While these methods provide a simple and effective option for human teachers, people are different. For each person and task there may be a different ideal feedback that properly balances ease and informativeness. \citet{baxter_coactive} highlights this by having users train a manipulator on a two tasks with two forms of feedback, ranking and physical re-orientation. They find the ratio of type of feedback the users use differs across both users and tasks. Our observation is thus, in order to have an algorithm that can learn from different teachers and different tasks, it needs to be able to learn from a variety of noisy feedback.
\begin{figure*}[!hb]
\centering
\includegraphics[width=\linewidth]{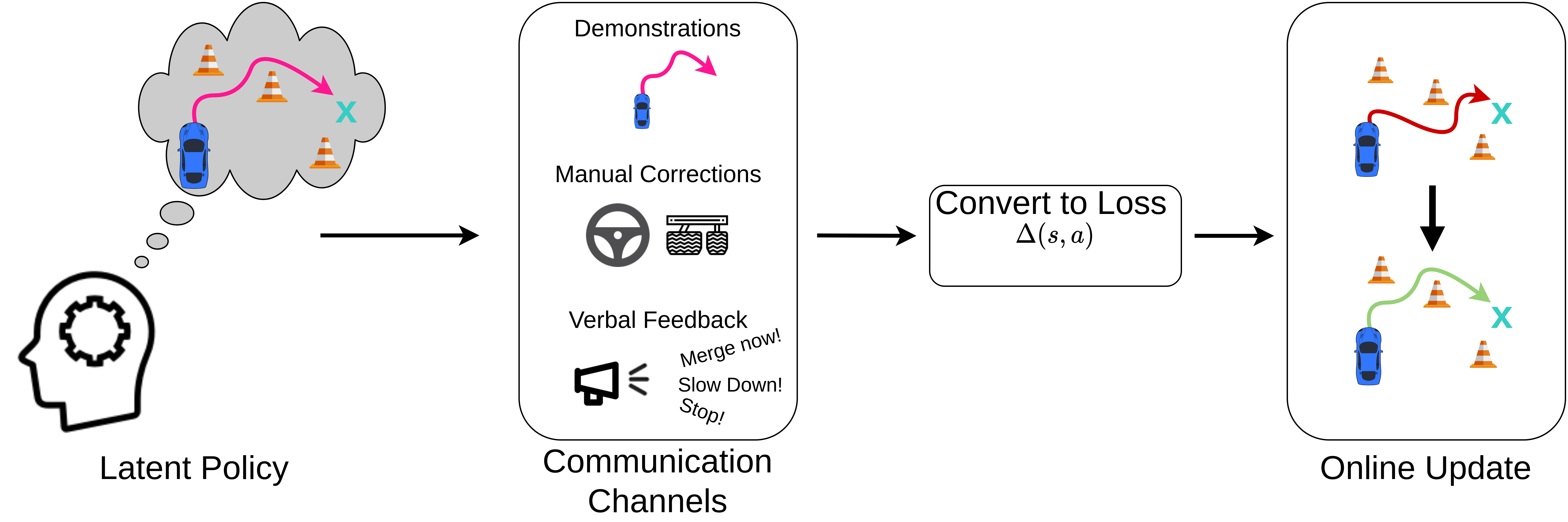}
\caption{The user has a desired latent policy where the car drives left through the cones to the blue X. The user can communicate their latent policy through a variety of noisy channels: a demonstration, online steering corrections, and/or verbal feedback to name a few. If we can model each channel with an associated loss function then we can update our autonomous policy through online learning.}
\label{fig:summary}
\end{figure*}

Building on the above observation we have the following key insight: The desired human policy is latent. Each feedback reveals some information about the latent policy. We can model a stream of feedback as a stream of loss functions. We wish to find a policy that has low loss on the sum of all loss functions. This naturally fits into the framework of online learning. This insight is illustrated in Fig. \ref{fig:summary}. Using this insight we create the Corrective Feedback Meta-Algorithm, which unifies prior work into an approach that can learn from a variety of noisy feedback. Specifically, our approach has the following properties:
\begin{enumerate}
    \item We can learn from a variety of feedback that needs only a mapping from feedback to loss.
    \item We can learn from noisy feedback.
    \item With few assumptions we can bound regret in terms of the teacher's latent loss.
    \item Prior work fits into our approach and benefits from our latent regret bounds.
\end{enumerate}
We demonstrate the our approach on a small scale autonomous racecar MuSHR~\citep{mushr} performing a household navigation task.
\begin{table}[t!]
\begin{center}
\begin{tabular}{ rcl  }
 \toprule 
 \multirow{2}{*}{Prior Work} &Noisy & \multirow{2}{*}{Feedback Type}\\
 & Feedback & \\
\midrule
\citet{Dagger}   & $\times$    & Actions\\
 \citet{coactive} & $\checkmark$  & Actions*\\
 \citet{Bajcsy:2017} & $\times$ & Trajectory Perturbations\\
 \citet{uncertainty_corrections} & $\checkmark$ & Trajectory Perturbations\\
 \citet{preference_learning} & $\checkmark$ & Binary Preferences\\
 \citet{coach}  & $\times$ &Reward\\
 \citet{deepcoach} & $\times$ &Reward\\
 \citet{good_bad_label}& $\checkmark$  & Binary Correct/Incorrect Feedback\\
 \citet{ARC}& $\checkmark$  & Good/Bad Labelling Parts of Trajectories\\
 \bottomrule
 \end{tabular}
\caption{Taxonomy of related works categorized by two characteristics: whether they model suboptimal or noisy feedback, and the feedback type. While many algorithms model noisy feedback, none explicitly allow for multiple forms of feedback. *Coactive Learning requires that the feedback can be converted into action level feedback.}
\label{table: taxonomy}
\end{center}
\end{table}
\section{Related Work}
Table \ref{table: taxonomy} shows related work as a taxonomy. Each approach is designed for a specific form of user feedback, and some approaches model user feedback as noisy. Below we give a brief summary of each approach. We also note that our approach does not fall under active learning~\citep{active_learning} as the robot never queries the teacher.   Instead, the teacher corrects when they deem necessary. 

DAgger~(\citet{Dagger}) collects action level feedback from the expert for every step and continually aggregates that feedback into its dataset for retraining. This is a very effective approach when the user can provide noiseless action level feedback. Additionally, it has been shown to be effective for robotics.

Coactive Learning~(\citet{coactive}) is very similar to our approach. We use a similar $\alpha$-informative noise model (see Section \ref{eq:alpah-inform}). Coactive learning has been used successfully in robotics~\citep{baxter_coactive, uncertainty_corrections}. We show in Section \ref{special cases} how Coactive Learning fits into our framework.

\citet{Bajcsy:2017} used trajectory perturbations to train a robotic manipulator to follow human preferred trajectories. Perturbations were done through physically pushing the manipulator. This work takes a Bayesian approach to updating the model parameters. They arrive at a similar update rule as ours and Coactive Learning.

\citet{uncertainty_corrections} builds on Coactive Learning by modeling uncertainty over the user's preferences via a Kalman filter. Similar to~\citet{Bajcsy:2017}, they use trajectory perturbations but differ in that corrections are not applied as the robot is executing, but after execution.

\citet{preference_learning} uses a form of preference learning~\citep{preference_intro} where they present two options to the user to decide between. From repeated queries, they build a reward function. In their feedback model, they assume that the human provides a random response 10\% of the time. In Section \ref{special cases}, We show how a simple version of this preference learning approach can fit into our framework.

COACH/DeepCOACH~(\citet{coach},~\citet{deepcoach}) ask the user for a reward signal and incorporate that reward using actor critic methods. \citet{good_bad_label} similarly use a simple correct/incorrect feedback in multi-class classification. These methods highlight how human preferred behaviors can be learned from simple feedback. 

\citet{ARC} takes an interesting approach to feedback and has the user label parts of the trajectory as good or bad. These trajectory critiques are used to create a reward function. They model noise by a confidence factor that represents how confident they can be in the user's feedback being accurate. They recommend adjusting this value based on how consistent the feedback is with previous feedback.

\section{Corrective Feedback Meta-Algorithm}
\begin{figure}[!htb]
\centering
    \textbf{Assumption 1}\\
    \includegraphics[width=0.7\linewidth]{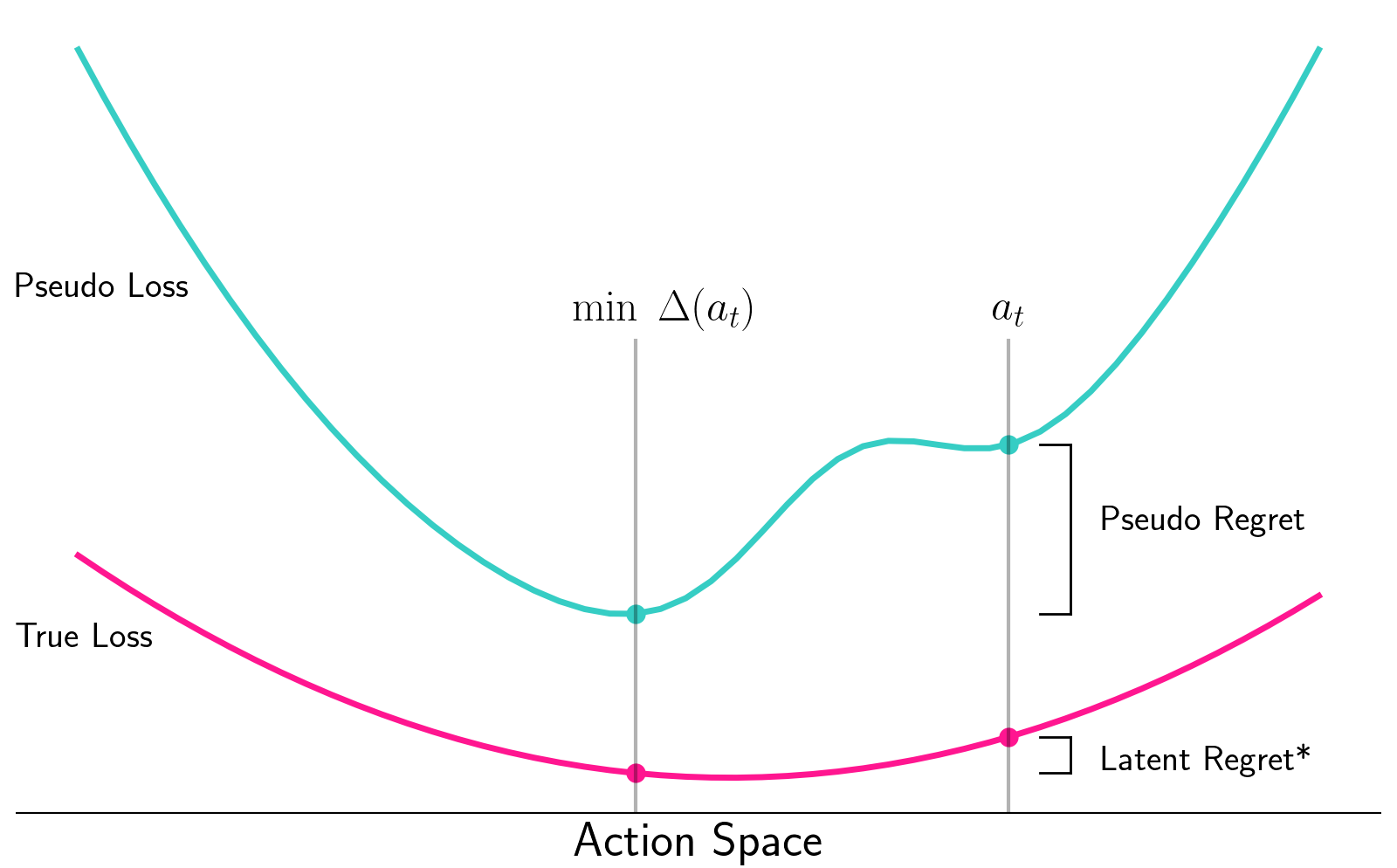}
    \caption{Our first assumption, the pseudo regret upperbounds the latent regret*. This is possible because we can make the pseudo loss have a higher curvature than the true loss. *Similar to the true latent regret except regret is relative to $\underset{a}{\argmin}\; \ploss_t(a)$ instead of $a^*$.}
    \label{fig:assump1}
\end{figure}
We present the Corrective Feedback Meta-Algorithm as an online interaction between a learner and a teacher. At round $t$, the learner observes the state $s_t \in \stateSpace$ and must select an action $a_t \in \actionSpace$. We model the teacher's preference via a loss function $\tloss_t : \actionSpace \rightarrow \mathbb{R}^{+}$ that assigns a cost to each action. If the learner had access to this cost function, they would select an optimal action $a_{t}^*$ via:
\begin{equation}
    a_{t}^* = \argmin_a \tloss_t(a)
\end{equation}

However, the teacher cannot provide $\tloss_t(a)$. Instead the teacher is compelled to provide feedback $z_t$ which we can convert into a \emph{pseudo-loss} $\ploss_t(a)$ which the learner can observe. The pseudo-loss is mismatched with the teacher's true loss - the question is how does this mismatch affect the regret bound w.r.t $\tloss$? We generalize the following assumptions from \cite{coactive} regarding $\ploss_t(a)$. 

We can assume the teacher's true loss is bounded and since we have flexibility in constructing the pseudo-loss using $z_t$ we can assume that the curvature is much higher than the true loss. With a higher curvature, the pseudo-regret upper bounds the regret relative to the minimum of the pseudo loss. This is seen visually in Fig. \ref{fig:assump1}.
\begin{equation}
    \label{eq:latent_pseudo}
    \underbrace{ \ploss_t(a_t) - \min_a \ploss_t( a ) }_{\text{pseudo regret}} \geq \underbrace{\tloss_t(a_t) - \tloss_t( \argmin_a \ploss_t(a) )}_{\text{learner versus the best pseudo action}}
\end{equation}

Additionally, we make the $\alpha$-informative assumption. It says - the best action according to the \emph{pseudo loss} may not be the best action according to the \emph{teacher's loss}, but is definitely a step in the right direction. 
\begin{equation}
    \label{eq:alpah-inform}
    \underbrace{\tloss_t(a_t) - \tloss_t( \argmin_a \ploss_t(a) )}_{\text{learner versus the best pseudo action }} \geq  \underbrace{\alpha \left( \tloss_t(a_t) - \tloss_t( a^*_t ) \right)}_{\text{learner versus the best latent action}}
\end{equation}

\begin{figure}[!htb]
\centering
    \textbf{Assumption 2}\\
    \includegraphics[width=0.8\linewidth]{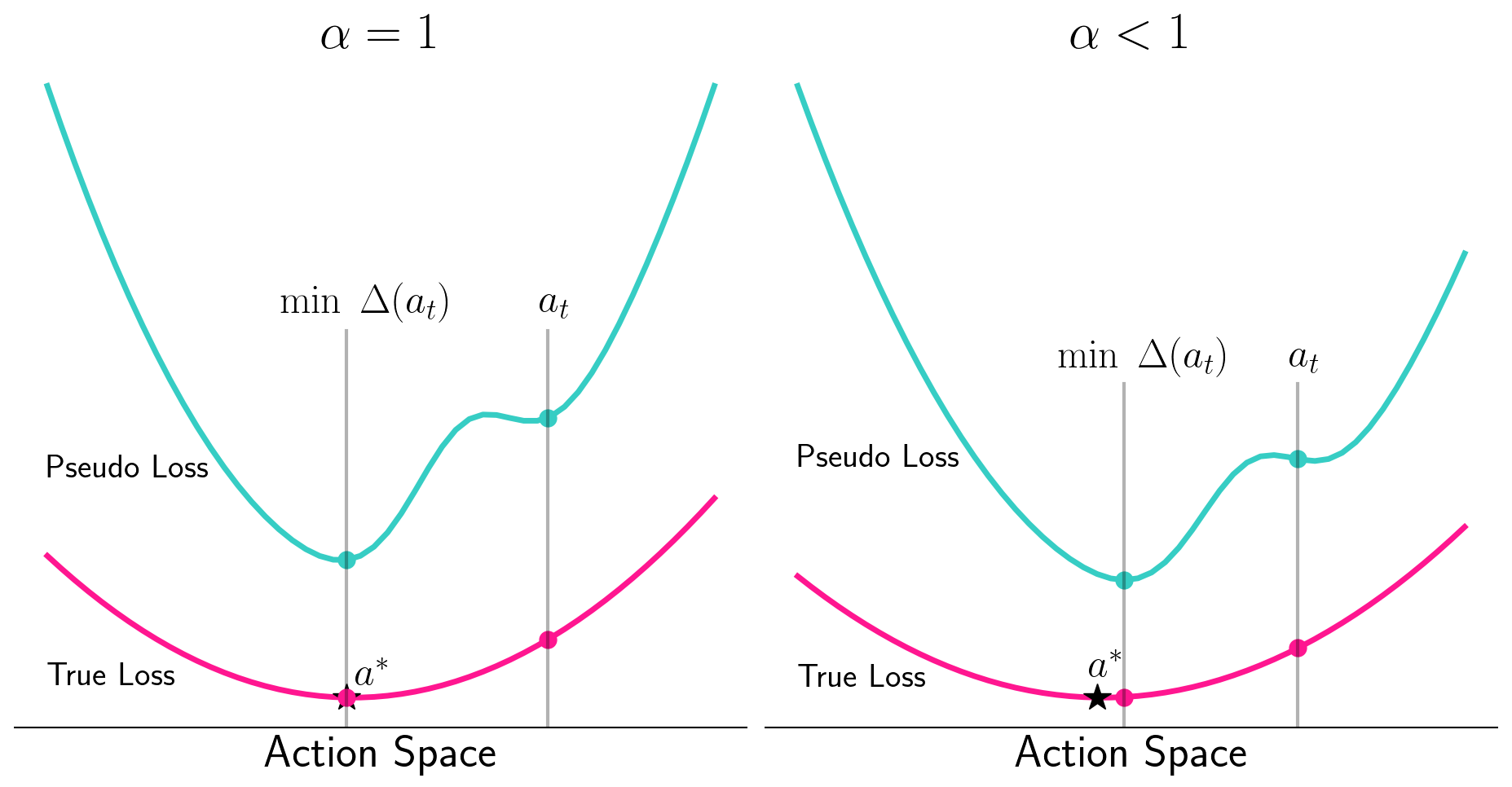}
    \caption{$\alpha$-informative assumption, the lower $\alpha$ is the more mis-aligned the pseudo loss is with the true loss.}
    \label{fig:assump2}
\end{figure}
$\alpha$ represents the amount of noise in the teacher's feedback, or more specifically how the mis-aligned the pseudo loss is with the true loss, see Fig. \ref{fig:assump2}. When $\alpha = 1$, the pseudo loss has no noise and the minimum of the pseudo loss is the minimum of the true loss. As $\alpha$ approaches 0, the pseudo loss becomes more and more mis-aligned with the true loss. Note, this parameter is purely for theoretical analysis and cannot be calculated in practice.

These assumptions place no requirements on the type of feedback, requiring only that it can be converted to a pseudo loss. We show in section \ref{theory} how these assumptions can be used to bound regret on the teacher's latent loss.
\newpage
\newtheorem{theorem}{Theorem}[section]
\section{Theoretical Analysis}\label{theory}
We bound latent regret using assumptions \ref{eq:latent_pseudo} and \ref{eq:alpah-inform}. Assume we have a policy class that maps weights $w$ and features $\phi_t(.)$ to actions, i.e., $a_t = \argmax_a \pi(\phi_t(a), w)$. We can transform the pseudo-loss over actions to a loss over weights
\begin{equation}
    \ploss_t(w) = \ploss_t(\argmax_a \pi(\phi_t(a), w))
\end{equation}

$\ploss_t(w)$ is non-convex and problematic to deal with. Hence, we derive a convex surrogate loss $\ell_t(w) \geq \ploss_t(w)$. Now we can bound regret via online learning. Here  are the steps:

\begin{enumerate}
    \item Use \emph{any online learning algorithm} to bound regret on surrogate loss
    \begin{equation}
    \begin{aligned}
        \sum_{t=1}^T \ell_t(w_t) - \min_{w^*} \sum_{t=1}^T \ell_t(w^*) \leq \mathcal{O}(\sqrt{T})\\
        \sum_{t=1}^T \ell_t(w_t) \leq \min_{w^*} \sum_{t=1}^T \ell_t(w^*) + \mathcal{O}(\sqrt{T})\\
    \end{aligned}
    \end{equation}
    Let's assume we have an expressive function class so $\min_{w^*} \sum_{t=1}^T \ell_t(w^*) = 0$ or at the most $\mathcal{O}(\sqrt{T})$.
    \item We can bound the pseudo loss on actions selected by the surrogate loss
    \begin{equation}
    \begin{aligned}
        \sum_{t=1}^T \ploss_t(a_t) = \sum_{t=1}^T \ploss_t(w_t) \leq \sum_{t=1}^T \ell_t(w_t) \leq \mathcal{O}(\sqrt{T})
    \end{aligned}
    \end{equation}
    For simplicity, we can zero bias the vector $\min_a \ploss_t( a ) = 0$.

    \item Combining the above with assumptions (\ref{eq:latent_pseudo}) and (\ref{eq:alpah-inform}), we get the following bound on true regret.
    \begin{equation}
    \begin{aligned}
    \label{eq:cum_regret}
        &\sum_{t=1}^T \alpha ( \tloss_t(a_t) - \tloss_t( a^* ) ) \leq \sum_{t=1}^T \ploss_t(a_t) \leq \mathcal{O}(\sqrt{T}) \\
%        & \leq \sum_{t=1}^T \tloss_t(a_t) - \tloss_t( \argmin{a} \ploss_t(a) )\\
        & \sum_{t=1}^T  ( \tloss_t(a_t) - \tloss_t( a^* ) ) \leq \frac{1}{\alpha} \mathcal{O}(\sqrt{T})
    \end{aligned}
    \end{equation}
\end{enumerate}

This can hold for any choice of latent loss, function classes, and surrogate functions that follow the assumptions (\ref{eq:latent_pseudo}) and (\ref{eq:alpah-inform}).
\section{Handling a Variety of Feedback}
We demonstrate how our approach works with arbitrary feedback, and how Coactive Learning and Preference Learning fit into our meta-algorithm. Note, the presented cases serve as examples and are not representative of all the function classes, feedback, and loss functions that can be handled within this approach.

\label{special cases}
\subsection{Linear Models \& Arbitrary Feedback}\label{generalized_hinge}
We first apply our approach to linear models with arbitrary feedback. We consider policies of the form:
\begin{align*}
    \pi_w(s) = \argmax_{a \in \mathcal{A}} w^{\intercal} \phi(s,a)
\end{align*}
where $\mathcal{A}$ is a discrete set of $k$ actions, $s$ is a state, $w$ is a weight vector, and $\phi$ is the feature vector. The policy may receive a correction $z_t \in \mathcal{Z}$ from the user. For example, Dagger~\citep{Dagger} considers corrections in the form of alternative actions, although we also consider more general forms of feedback. The feedback is indicative of the quality of various actions. Recall the pseudo-loss: $\Delta_t(a)$. For the remainder of this section, we will rewrite this to depend on the current state instead of time $\Delta(s, a)$. To incorporate $z_t$, we will depend on the following \emph{user model}:
\begin{enumerate}
    \item If the user explicitly provides a correction $z$, we define $\Delta(s,a) = \xi(a,z)$ where $\xi$ is a user-defined similarity function between the action and feedback spaces. 
    \item If the action $a_t$ is ``good enough'', the user does not correct, i.e.\ $z=\varnothing$. We will model this by not applying a loss.
\end{enumerate}
When feedback is given, we have a vector of losses $\Delta(s,\cdot)$ for all actions. We upper bound this pseudo loss with the following surrogate convex loss over weights called the \emph{generalized hinge loss}~\citep{understanding_ml}.
\begin{equation}
\label{eq:generalized_hinge}
    \ell(w) = \max_{i = 1, \dots, k} \Delta(s, a_i) + w^{\intercal} (\phi(s, \hat{a}) - \phi(s, a_i))
\end{equation}
where $\hat{a} = \argmin_{a \in \mathcal{A}} \Delta(s,a)$ or the best action according to the pseudo-loss. The subgradient of the loss is
\begin{equation}
    \nabla_w \ell(w) =  \phi(s,\hat{a}) - \phi(s, a_{\text{sel}})
\end{equation}
Where $a_{\text{sel}}$ is the action that achieves the maximum in (\ref{eq:generalized_hinge}). 
Since this fits into the our approach, we enjoy the same regret bounds as in equation (\ref{eq:cum_regret}).
\subsection{Coactive Learning}
We show that Coactive Learning also fits into our approach. Specifically, we are looking at the Preference Perceptron algorithm from~\citet{coactive} not considering the slack variable. The perceptron policy class is defined as follows:
\begin{equation}
    \pi_w(s) = \argmax_{a \in \mathcal{A}} w^{\intercal} \phi(s,a)
\end{equation}
They use a latent utility function that they are trying to maximize:
\begin{equation}
    \tloss(s, a) = w_*^T\phi(s,a)
\end{equation}
where $w_*$ is the unknown teacher weights. Note, $\pi_{w_*}(s) = \argmax_{a \in \mathcal{A}}\tloss(s, a)$. Since the teacher's actions (feedback) are a noisy representation of this policy, we can define a pseudo 0-1 loss that we will minimize. 
\begin{equation}
    \Delta(s,a) = \mathbb{I}(a \neq \hat{a})
\end{equation}
This upper bounds the negated objective: $\pi_{w_*}(s) = \argmin_{a \in \mathcal{A}}-\tloss(s, a)$ where $\hat{a}$ is the expert provided action. Finally, we can upper bound this loss with a convex surrogate.
\begin{equation}
    \Delta(s,a) \leq \ell(w) = 1 + w^{\intercal} (\phi(s, a) - \phi(s, \hat{a})))
\end{equation}
Note, this breaks our assumption $\min_{w^*} \sum_{t=1}^T \ell_t(w^*) \leq \mathcal{O}(\sqrt{T})$ and is now bounded by $\mathcal{O}(T)$. This produces a latent regret bound of the following:
\begin{equation}
\sum_{t=1}^T  ( \tloss_t(a_t) - \tloss_t( a^* ) ) \leq \frac{1}{\alpha} \mathcal{O}(T + \sqrt{T})
\end{equation}
We can take advantage of this being a linear policy to tighten this vacuous bound. If we take the derivative of $\ell_t(w)$ with respect to $w$, we obtain the Coactive update rule.
\begin{equation}
    w_{t+1} = w_t + \phi(s, \hat{a}) - \phi(s, a)
\end{equation}
Then, we can follow the Coactive Learning proof to obtain their regret bounds, which are the same worst case as our latent regret, $\mathcal{O}(\sqrt{T})$.

\subsection{Preference Learning}
We show that preference learning also fits into our approach. We focus on the simplest case, where the teacher is presented with  two actions: the learner's choice, and another random choice. The teacher must choose which action they prefer. Given a teacher's choice, we can construct a pseudo-loss as follows:
\begin{equation}\label{eq:preference}
    \Delta(s,a) = \norm{\phi(s, a_p) - \phi(s, a)} - \norm{\phi(s, a_{np}) - \phi(s,a)}
\end{equation}
where $a_p$ is the preferred action and $a_{np}$ is the not-preferred action. This promotes actions that are similar to $a_p$ and simultaneously dissimilar from $a_{np}$. Using this pseudo loss, we can use the generalized hinge loss as in \ref{generalized_hinge} and recover the same regret bounds.

\begin{figure}[!htb]
\minipage{0.32\textwidth}
  \centering
  \includegraphics[width=\linewidth]{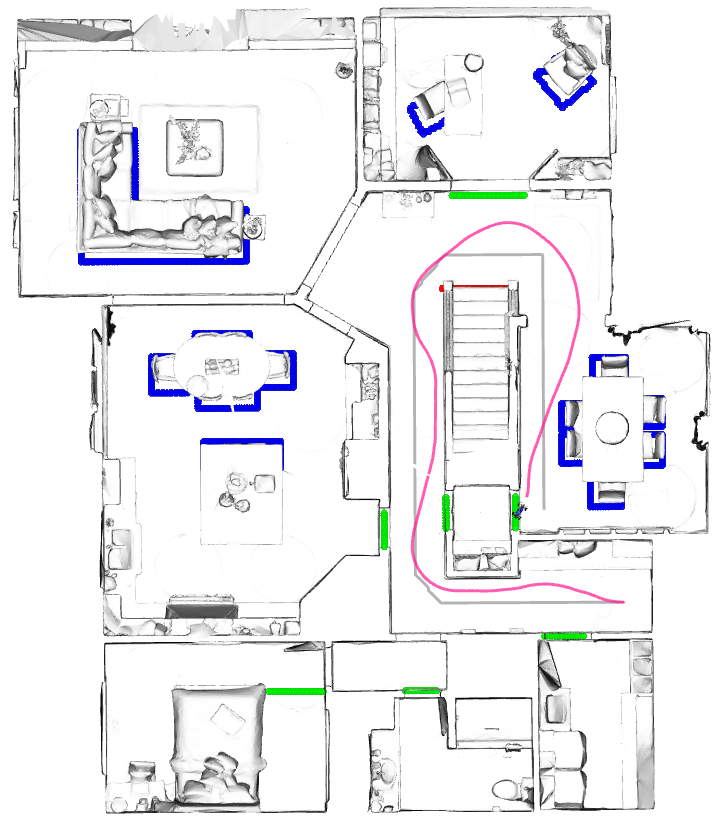}
  \\ Avoid Stairs $\vert$ Tolstoy
\endminipage\hfill
\minipage{0.32\textwidth}
  \centering
  \includegraphics[width=\linewidth]{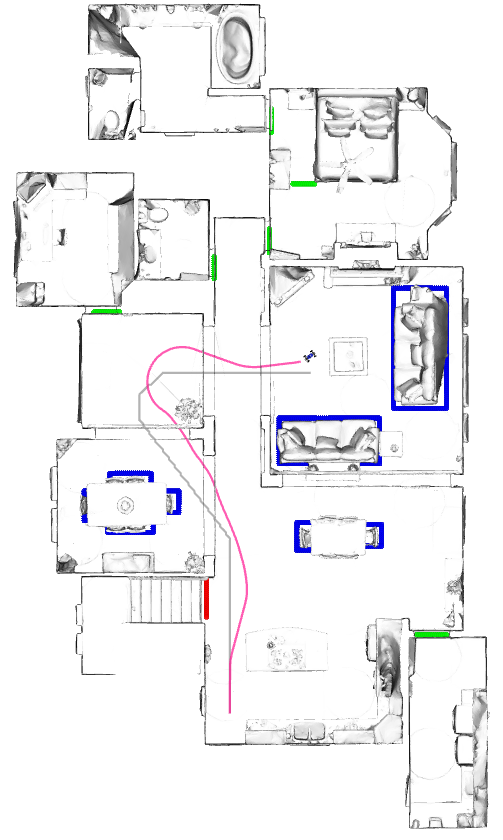}
  \\ Avoid Stairs \& Chairs $\vert$ Corozal
\endminipage\hfill
\minipage{0.32\textwidth}%
  \centering
  \includegraphics[width=\linewidth]{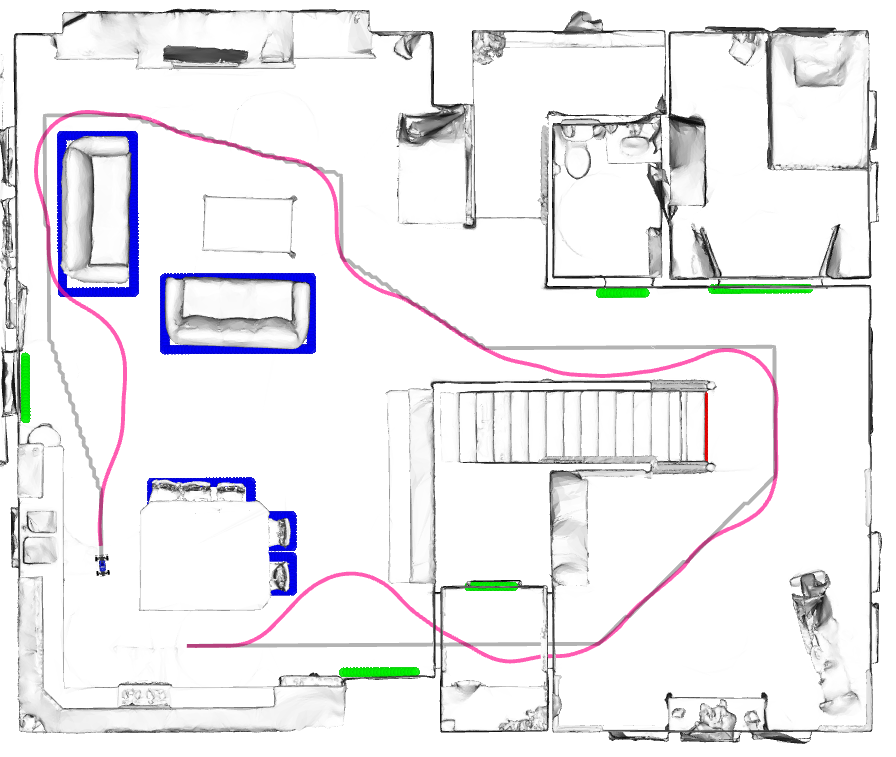}
  \\ Avoid Doors $\vert$ Newfields
\endminipage\\
\centering
\includegraphics[width=0.6\linewidth]{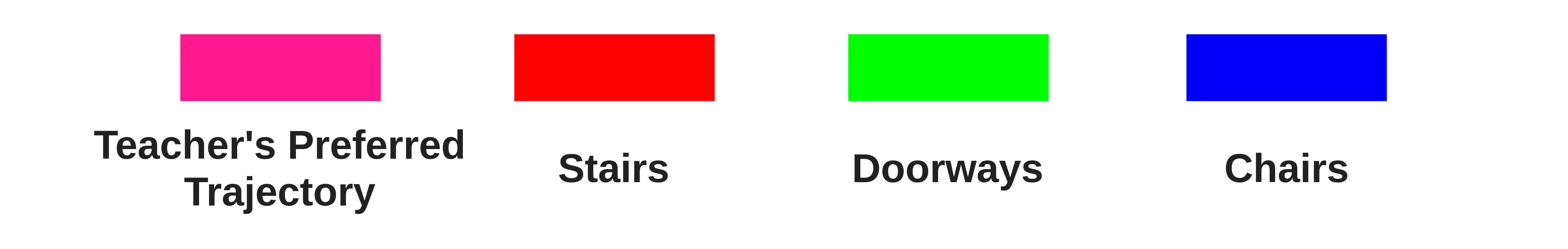}
\caption{Our semantically labelled testing environments and teacher trajectory. The gray line is the fixed path that the robot should follow.}
\label{fig:teachers}
\end{figure}

\section{Experimental Evaluation}
To demonstrate our approach we perform simulated household navigation experiments with the MuSHR Racecar Platform~\citep{mushr}. The task is to follow fixed paths throughout various households per the teacher's preferences. At each timestep, the learner must choose from 64 actions represented as 1 meter long trajectories from which a model predictive controller will calculate a steering angle and velocity. For each trajectory, the feature function for our linear policy $\phi(s, a)$ provides:
\begin{itemize}
    \item Sum of distances to obstacles along the trajectory.
    \item Sum of distances to semantically labelled objects: stairs, chairs, and doors.
    \item Sum of cross track and along track error for the trajectory relative to a reference point on the path 1 meter ahead.
    \item Relative lateral displacement from end of trajectory to robot. Trajectories going straight are 0, ones to the right are positive, and ones to the left are negative.
\end{itemize}
We use 3 environments from the Gibson Simulator~\citep{xiazamirhe2018gibsonenv}: Tolstoy, Newfields, and Corozal. While we don't allow the robot to choose trajectories in collision , if the robot gets into an unrecoverable state and collides, then we reset it to the starting point maintaining its current weights.

We train from three programmatic teachers, one that wants to avoid doors, another that wants to avoid stairs and stay to the right of the path, and another that wants to avoid chairs and stairs. We choose these behaviors because we expect a user would want the robot to avoid places from which people could suddenly move from and lateral position is important for hallways where it is convention to stay to a particular side. Programmatic teachers allow us to calculate latent loss as a metric for learning. All teachers also want the robot to follow the path provided. See Fig. \ref{fig:teachers} for each map and teacher's desired behavior. 

We use several forms of feedback. We use online gradient descent to update our linear policy and note that other optimization methods could learn with fewer corrections. For all forms of feedback, we upperbound each pseudo loss with the generalized hinge loss. Specifically, we use:
\begin{itemize}
    \item \textbf{Action:} Feedback is the desired teacher action from the 64 potential actions. We use a pseudo loss where the teacher's action is 0 and the others are 100. 
    \item \textbf{Preference:} The teacher is presented with two actions: the learner's choice and another random action. They choose which action they prefer more. We use the pseudo loss described in Equation \ref{eq:preference}.
    \item \textbf{Semantic:} This rough form of feedback is intended to simulate a teacher telling the robot to ``avoid doors" etc. Given the teacher's and learner's actions, we determine if the teacher's action features prefer/avoid an action closer/further from semantic objects as compared to the learner. If they prefer the object, then all actions that prefer the object more than the learner's choice are weighted 0 and all others 100 (visa-Versa for avoiding objects). We also include the path (via cross track error) as a semantic object to represent the teacher saying ``stay on path". Our pseudo loss is the sum of all the semantic losses (doors, stairs, chairs, and path).
\end{itemize}
\begin{figure}[!htb]
\centering
\textbf{We Learn From a Variety of Feedback}\par\medskip
\minipage{0.33\textwidth}
  \includegraphics[width=\linewidth]{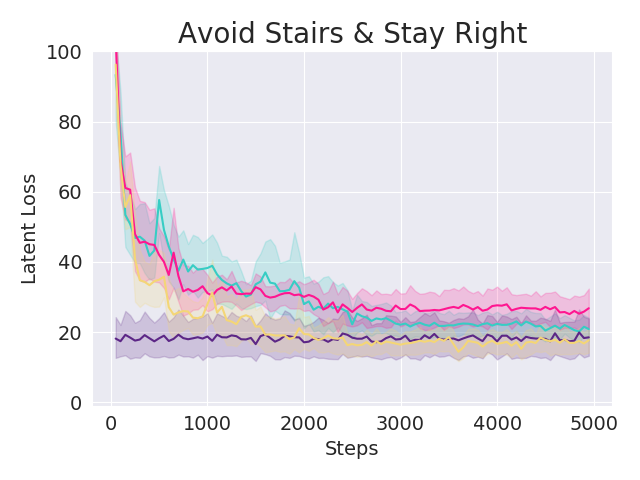}
\endminipage\hfill
\minipage{0.33\textwidth}
  \includegraphics[width=\linewidth]{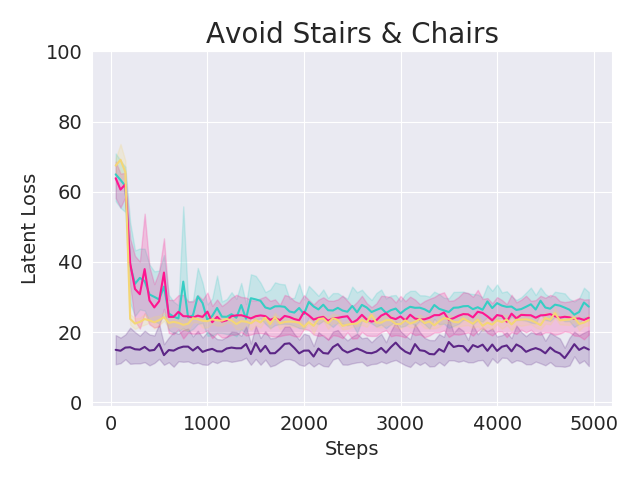}
\endminipage\hfill
\minipage{0.33\textwidth}%
  \includegraphics[width=\linewidth]{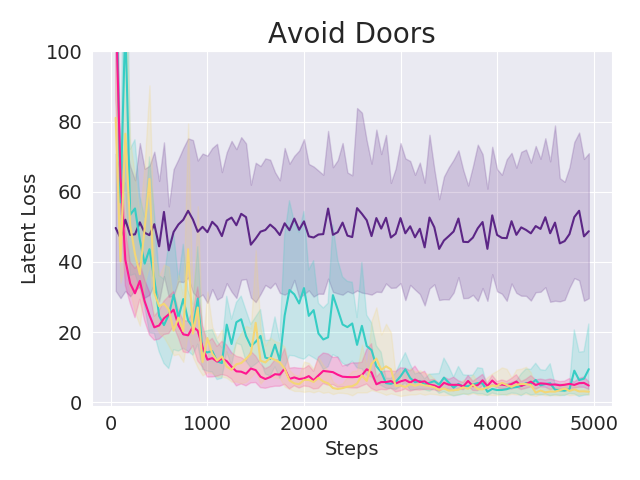}
\endminipage\\
\minipage{0.33\textwidth}
  \includegraphics[width=\linewidth]{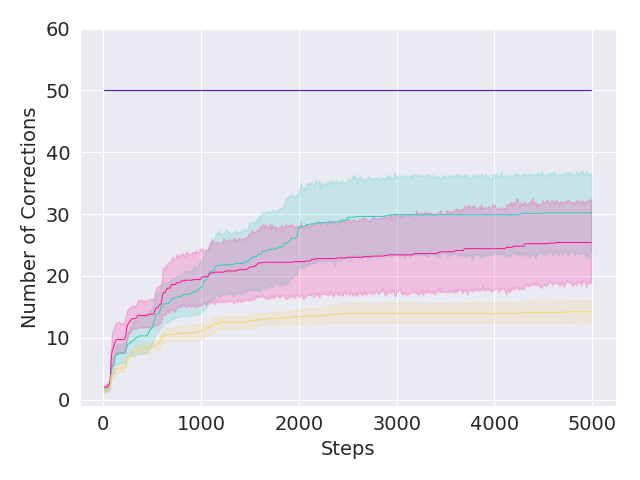}
\endminipage\hfill
\minipage{0.33\textwidth}
  \includegraphics[width=\linewidth]{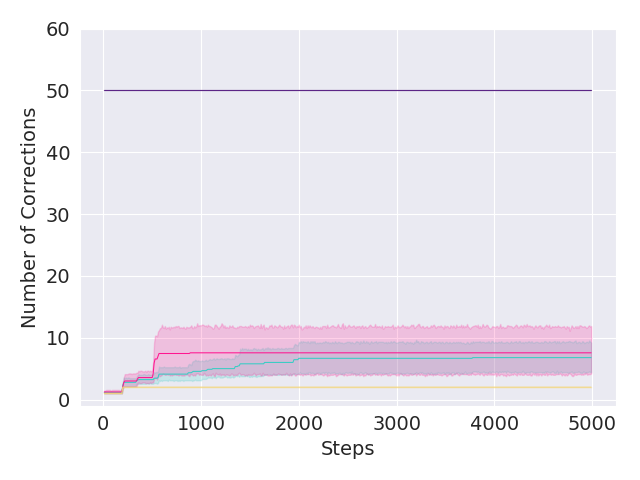}
\endminipage\hfill
\minipage{0.33\textwidth}%
  \includegraphics[width=\linewidth]{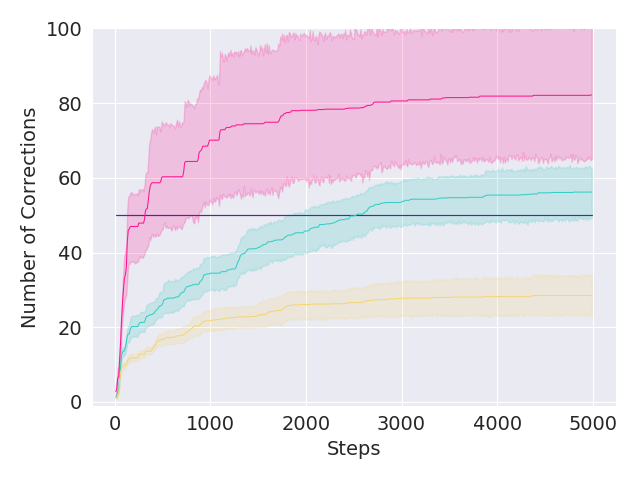}
\endminipage
\\
\centering
\includegraphics[width=0.4\linewidth]{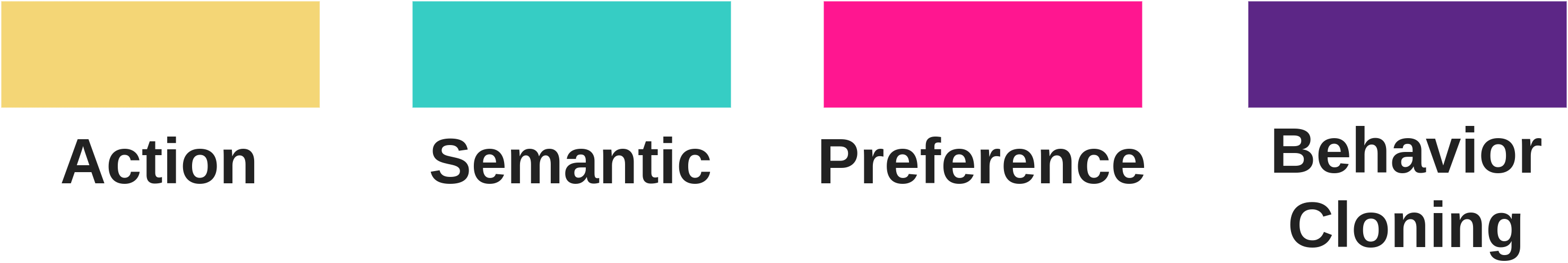}
\caption{Top Row: Latent loss over time. Note how all policies converge to a low loss except for Avoid Stairs \& Chairs. This is due to the teacher threshold for correcting being too high. Bottom Row: Number of corrections over time. Note how the order of convergence is different across tasks.}
\label{fig:variety}
\end{figure}
\subsection{Learning from a Variety of Feedback}
To demonstrate how our approach can learn from a variety of corrective feedback, we test each form of feedback in three environments, where each environment employs a different teacher as shown in Fig. \ref{fig:teachers}. For comparison, we also run behavior cloning trained on 50 samples of action feedback. Our metrics for learning are the latent loss: $w^T_*(\phi(s,a) - \phi(s,a^*))$ and the number of corrections. The teacher corrects whenever the latent loss is above 1.0. The value of this threshold translates to how picky the teacher is. We investigate how setting this parameter affects convergence in Section \ref{thresholds}. For each feedback in each environment, we conduct 10 trials. We use a learning rate of 0.01. As learning progresses, we expected to see the latent loss decline and the number of corrections to stop growing. Results are shown in Fig. \ref{fig:variety}.

The first row of the results show the latent loss over time. In Avoid Doors, all methods of feedback converge near zero in 5000 steps. Behavior cloning remains constant because it does not receive additional training from the initial first 50 datapoints. In Avoid Stairs \& Stay Right and Avoid Stairs \& Chairs, we see that all methods converge with a latent loss around 20. This is due to the teacher threshold being too high for the task. We know the threshold is too high because below the latent loss plot, we see the number of corrections over time flattening out, indicating that the teacher is satisfied. Without corrections, these models cannot improve.

Looking at the number of corrections over time (second row), we see that action feedback converges with the least number of corrections consistently, suggesting this is the most informative form of feedback. We also note that each task requires a different number of corrections, because the path and environment are different between each. Interestingly, semantic and preference feedback do not have a clear order across tasks. This finding shows that a particular form of feedback is not always the best choice across tasks and environments highlighting the need for a variety of feedback types.

\subsection{Learning from Noisy Feedback}
\begin{figure}[!htb]
\centering
\textbf{We Learn Under Noisy Feedback}\par\medskip
    \minipage{0.5\textwidth}
      \includegraphics[width=\linewidth]{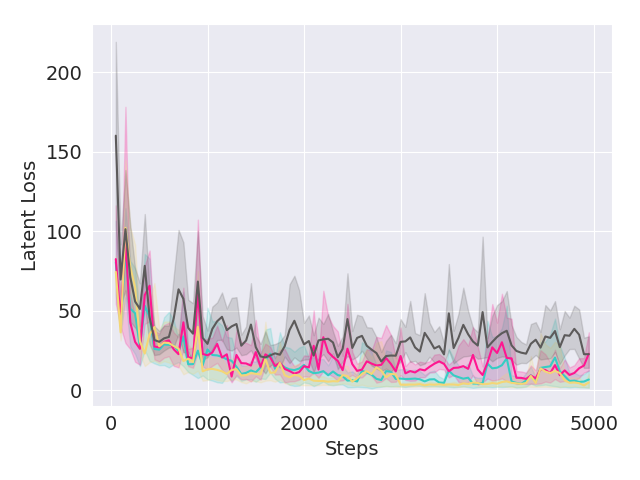}
    \endminipage\hfill
    \minipage{0.5\textwidth}%
      \includegraphics[width=\linewidth]{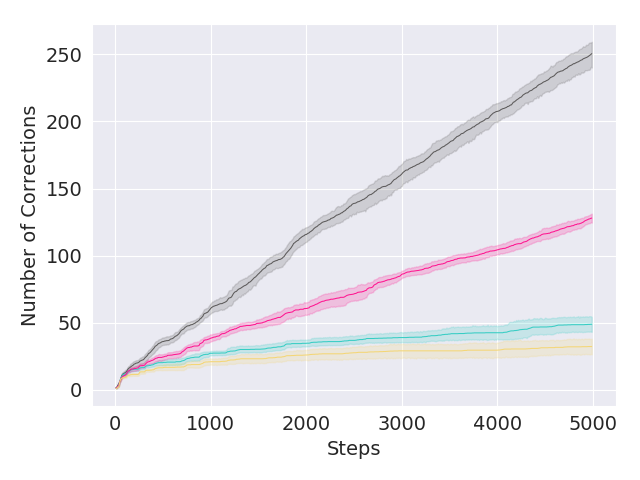}
    \endminipage\\
    \centering
    \includegraphics[width=0.25\linewidth]{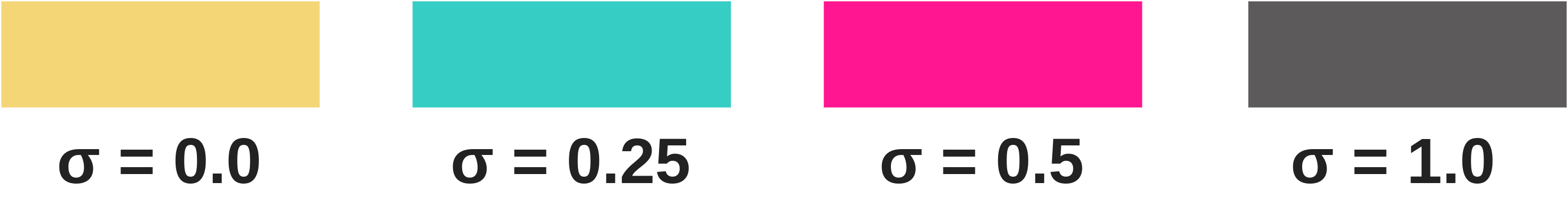}
    \caption{Left is the latent loss over time for four different degrees of noise and right is the associated number of corrections. Notice how with more noise, more corrections are supplied.}
    \label{fig:noise}
\end{figure}
\begin{figure}[!htb]
    \centering
    \minipage{0.33\textwidth}
      \centering
      \includegraphics[width=\linewidth]{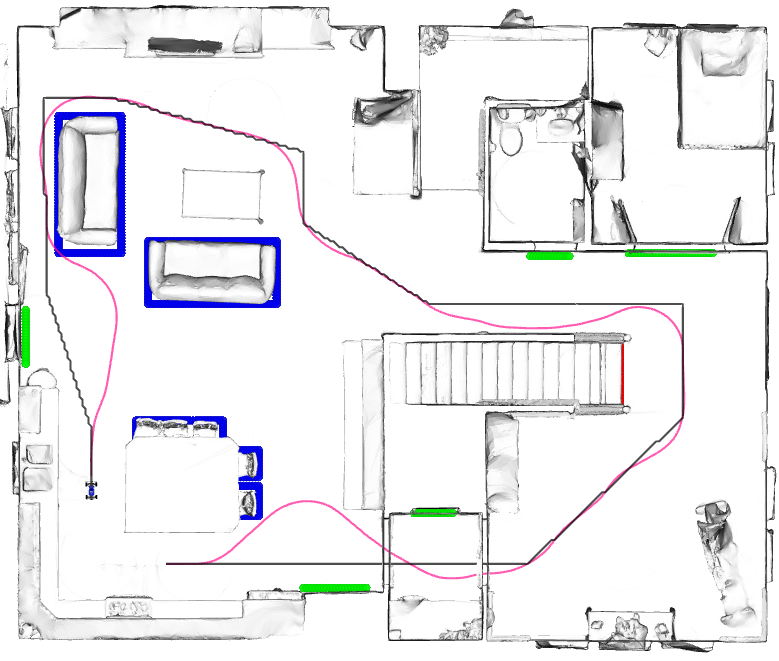}
      \\ 0.0
    \endminipage\hfill
    \minipage{0.33\textwidth}%
      \centering
      \includegraphics[width=\linewidth]{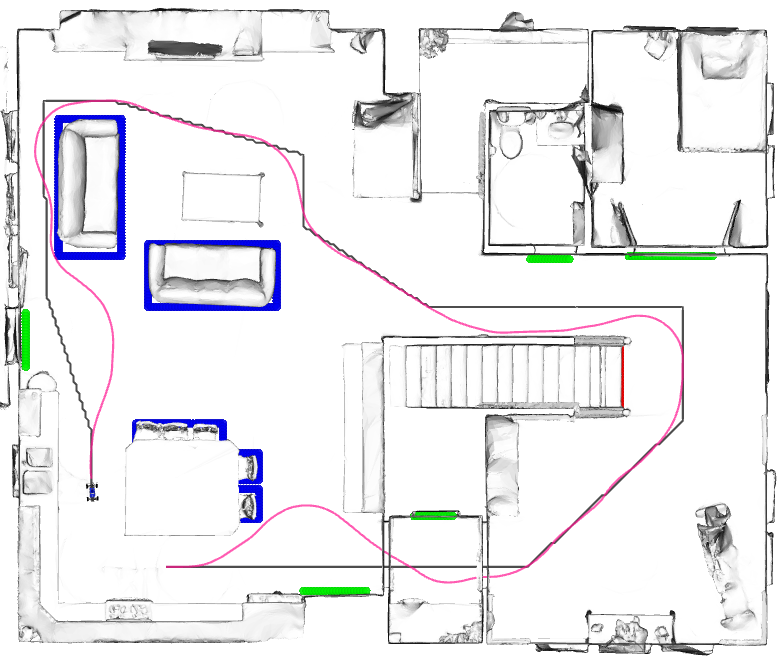}
      \\ 1.0
    \endminipage\hfill
    \minipage{0.33\textwidth}%
      \centering
      \includegraphics[width=\linewidth]{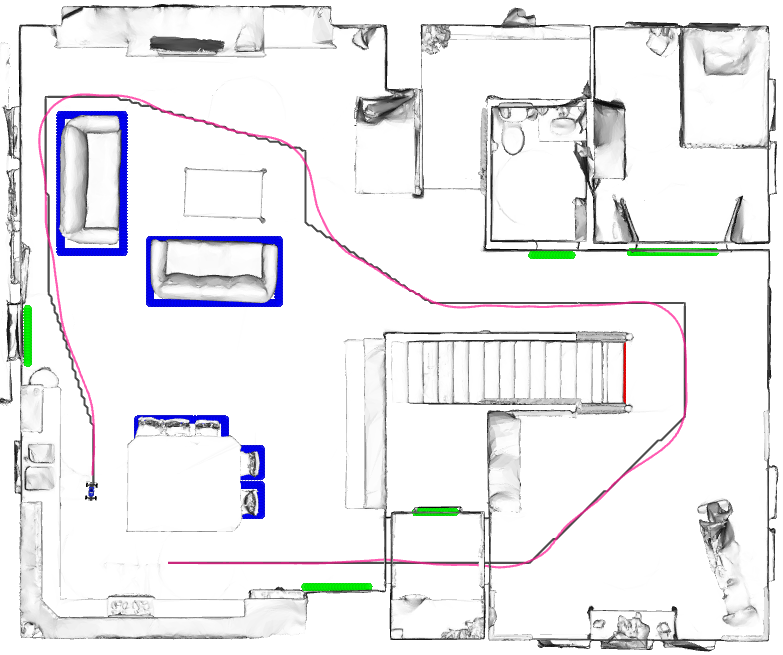}
      \\ 2.0
    \endminipage\hfill
    \caption{Paths taken after training with three different standard deviations of noise. Little difference in the quality of the run is seen between $\sigma = 0.0$ and $\sigma = 1.0$ but at $\sigma = 2.0$ the learner takes an alternate path to the end, not following the task.}
    \label{fig:noise_qual}
\end{figure}
To test our approach under noisy teacher feedback, we performed an experiment where we add Gaussian noise to the teacher's feedback. To calculate each form of feedback, we convert the latent loss into $\Delta(s,a)$ as described above. We add Gaussian noise to the latent loss, which creates both noisy feedback and a noisy frequency of corrections. We test with four standard deviations, 0.0 or no noise, 0.25, 0.5, and 1.0. We ran 10 trials on the Newfields environment with the teacher Avoid Doors and a latent threshold of 1. We used action feedback for this experiment.

As shown on the left in Fig. \ref{fig:noise}, our approach learns from noisy feedback. Unsurprisingly with more noise, the latent loss will increase. To get a sense of how the latent loss scales with more noise we performed a second experiment in the same environment where we increased the noise and measured the latent loss after 5000 steps. On the right in Fig. \ref{fig:noise} we see how more corrections are given with more noise. This is because noisier corrections provide the learner with a less accurate loss to update from. To give context to these loss values we provide paths taken by learner's after training for three different noise values in Fig. \ref{fig:noise_qual}. While the latent loss between $\sigma = 0.0$ and $\sigma = 1.0$ is different, qualitatively the learner still does a good job at the task with $\sigma = 1.0$. At $\sigma = 2.0$ the learner does not care for the doors at all, but simply follows the path.

\subsection{Effect of Different Teacher Thresholds}\label{thresholds}
\begin{figure}[!htb]
    \centering
    \textbf{Different Thresholds Result in Different Behaviors}\par\medskip
    \minipage{0.25\textwidth}
      \centering
      \includegraphics[width=\linewidth]{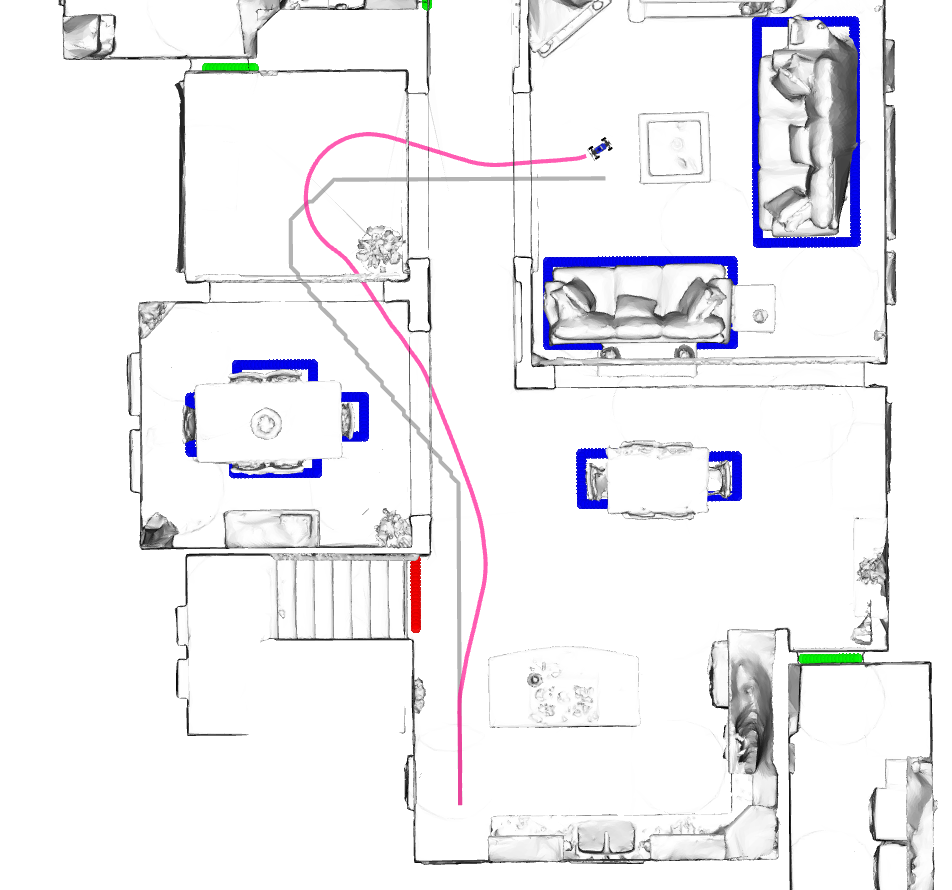}
       \\ 0.0
    \endminipage\hfill
    \minipage{0.25\textwidth}%
      \centering
      \includegraphics[width=\linewidth]{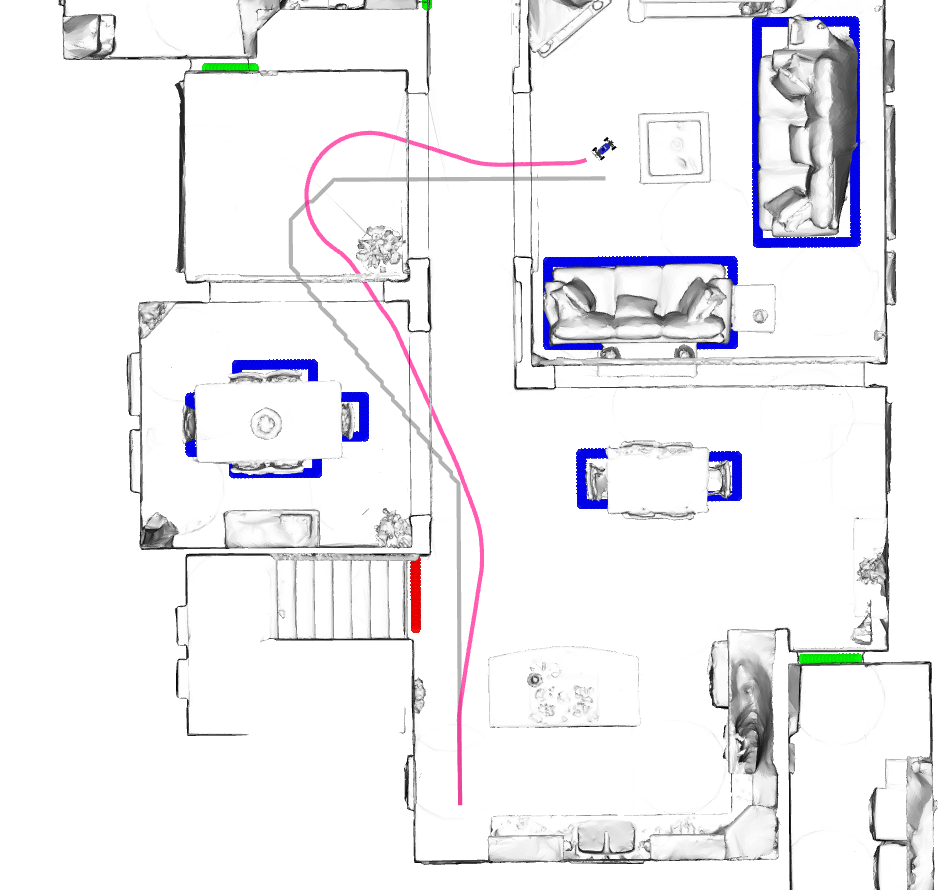}
       \\ 0.25
    \endminipage\hfill
    \minipage{0.25\textwidth}
      \centering
      \includegraphics[width=\linewidth]{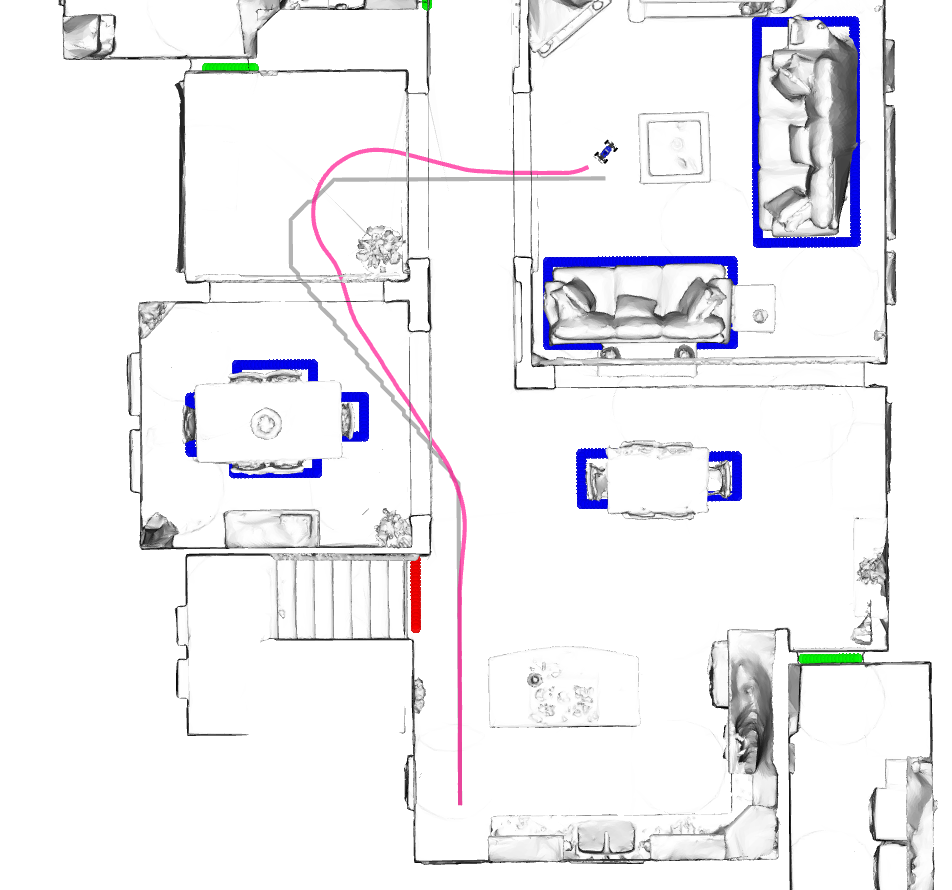}
       \\ 0.5
    \endminipage\hfill
    \minipage{0.25\textwidth}
      \centering
      \includegraphics[width=\linewidth]{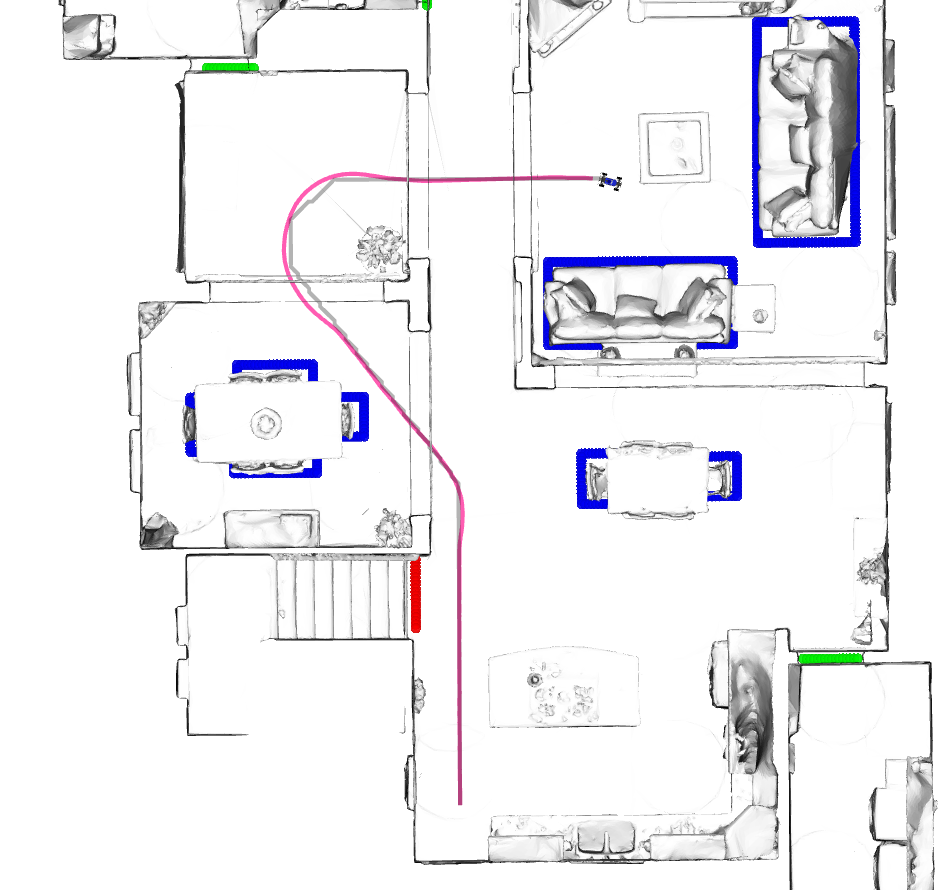}
        \\ 1.0
    \endminipage
    \caption{Learner's paths given different correction thresholds. From left to right, as the threshold increases the behavior gets further from the desired behavior avoid stairs and doors.}
    \label{fig:thresholds_qual}
\end{figure}

\begin{figure}[!htb]
    \centering
    \textbf{Lower Thresholds Result in Lower Loss \& More Corrections}\par\medskip
    \minipage{0.5\textwidth}
      \centering
      \includegraphics[width=\linewidth]{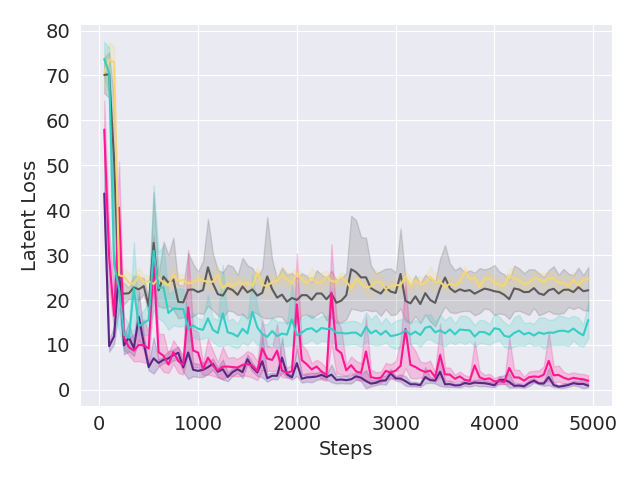}
    \endminipage\hfill
    \minipage{0.5\textwidth}%
      \centering
      \includegraphics[width=\linewidth]{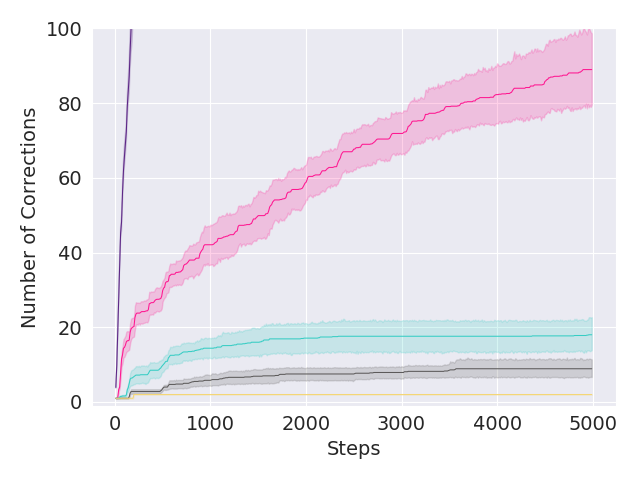}
    \endminipage\hfill\\
    \centering
    \includegraphics[width=0.5\linewidth]{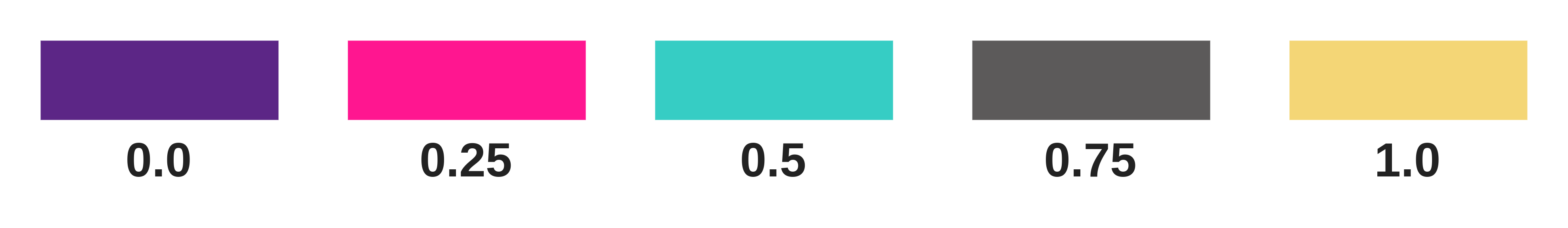}
    \caption{Left: Latent loss for Avoid Stairs \& Chairs for different latent thresholds. Right: The corresponding number of corrections over time for each threshold.}
    \label{fig:thresholds_quant}
\end{figure}

A human teacher's threshold for correcting is unknown and different for each task. Thus, we wanted to explore how different numerical thresholds affect the results. We test 5 thresholds: 0.0, 0.25, 0.5, 0.75, and 1.0. A threshold of 0.0 means that the teacher will correct every time the learner chooses a different action than the teacher's. In contrast, a higher threshold can result in the learner not converging to the desired policy. We test these thresholds in the Corozal environment with the Avoid Stairs \& Chairs teacher. We run each threshold for 10 trials and present quantitative and qualitative results for each in Fig. \ref{fig:thresholds_qual} and Fig. \ref{fig:thresholds_quant}.


We see qualitatively that lower thresholds result in a behavior most similar to the teacher's desired behavior from Fig. \ref{fig:teachers}. As the threshold increases, the learner does not care to avoid the stairs or the chair and just focuses on following the line. Quantitatively, we see a similar trend, lower thresholds result in lower loss and more corrections. Ideally, a human corrector will want to be in the middle ground, in our case, around 0.25 - 0.5 so they can recover their desired policy while also not having to make too many corrections.
\section{Discussion \& Future Work}
In this work, we present the Corrective Feedback Meta-Algorithm, which can learn from a variety of noisy feedback. For different tasks, teachers, and environments, different forms of feedback will be necessary. This work builds on prior work and creates a straightforward approach to encompass a variety of feedback. Our approach is built off of the insight that the teacher's policy is latent and that their feedback can be modelled as a stream of loss functions. We demonstrated this approach on three simulated household navigation tasks with different teachers in each. Our findings show that our method learns from both a variety of feedback and noisy feedback. While our method can learn successfully, we also highlight that teacher's who provide more feedback obtain policies with lower loss.

We see two interesting directions for future work. First is to apply this approach to a real world robotics scenario, with human teachers providing corrections. This would provide insights regarding when user's correct, and what types of corrections users use. Second, we would like to investigate how we can learn mappings from feedback to pseudo loss functions from prior experience instead of encoding each feedback type manually. This would significantly scale the variety of feedback the teacher could provide.

% The maximum paper length is 8 pages excluding references and acknowledgements, and 10 pages including references and acknowledgements

%\clearpage
% The acknowledgments are automatically included only in the final version of the paper.
\acknowledgments{This work was (partially) funded by the National Institute of Health R01 (\#R01EB019335), National Science Foundation CPS (\#1544797), National Science Foundation NRI (\#1637748), the Office of Naval Research, the RCTA, Amazon, and Honda Research Institute USA.}

%===============================================================================

\bibliography{references}  % .bib
\end{document}